\documentclass[twocolumn]{article}
\usepackage{epsfig}
\usepackage{wcci2002}

\newcommand{\reinf}{{\sc reinforce}}

\newcommand{\commentout}[1]{}

\newcommand{\tuple}[1]{\langle #1 \rangle}

\title{Reinforcement Learning for Adaptive Routing}
\author{Leonid Peshkin ({\it pesha@ai.mit.edu}) \hspace{2.4cm} 
Virginia Savova ({\it savova@jhu.edu}) \\
MIT Artificial Intelligence Lab. \hspace{3.3cm}  Johns Hopkins University \\
Cambridge, MA 02139  \hspace{4.5cm}  Baltimore, MD 21218 \\
}

\begin{document}
\maketitle
\thispagestyle{empty}\pagestyle{empty}

\begin{abstract}
Reinforcement learning means learning a policy---a mapping of observations
into actions---based on feedback from the environment. The learning can be
viewed as browsing a set of policies while evaluating them by trial through
interaction with the environment. We present an application of gradient ascent
algorithm for reinforcement learning to a complex domain of packet routing in
network communication and compare the performance of this algorithm to other
routing methods on a benchmark problem.
\end{abstract}

%\thanks{Errata at {\tt www.ai.mit.edu/\~{}pesha/papers.html}} \\
\footnotetext{Errata at {\tt www.ai.mit.edu/\~{}pesha/papers.html}} 

\section{Introduction}
\label{intro}
Successful telecommunication requires efficient resource allocation
that can be achieved by developing adaptive control
policies. Reinforcement learning ({\sc
rl})~\cite{Kaelbling96,Sutton98a} presents a natural framework for the
development of such policies by trial and error in the process of
interaction with the environment. In this work we apply the {\sc rl}
algorithm to network routing. Effective network routing means
selecting the optimal communication paths. It can be modeled as a
multi-agent {\sc rl} problem. In a sense, learning the optimal control
for network routing could be thought of as learning in some
traditional for {\sc rl} episodic task, like maze searching or pole
balancing, but repeating trials many times in parallel with
interaction among trials.

Under this interpretation, an individual router is an agent which
makes its routing decisions according to an individual policy. The
parameters of this policy are adjusted according to some measure of
the global performance of the network, while control is determined by
local observations. Nodes do not have any information regarding the
topology of network or their position in it. The initialization of
each node, as well as the learning algorithm it follows, are identical
to that of every other node and independent of the structure of the
network. There is no notion of orientation in space or other
semantics of actions. Our approach allows us to update the local
policies while avoiding the necessity for centralized control or
global knowledge of the networks structure. The only global
information required by the learning algorithm is the network utility
expressed as a reward signal distributed once in an epoch and
dependent on the average routing time. This learning multi-agent
system is biologically plausible and could be thought of as neural
network in which each neuron only performs simple computations based
on locally available quantities~\cite{Savova02}.

\section{Domain} 

\begin{figure*}[!hbt]
\centerline{\psfig{file=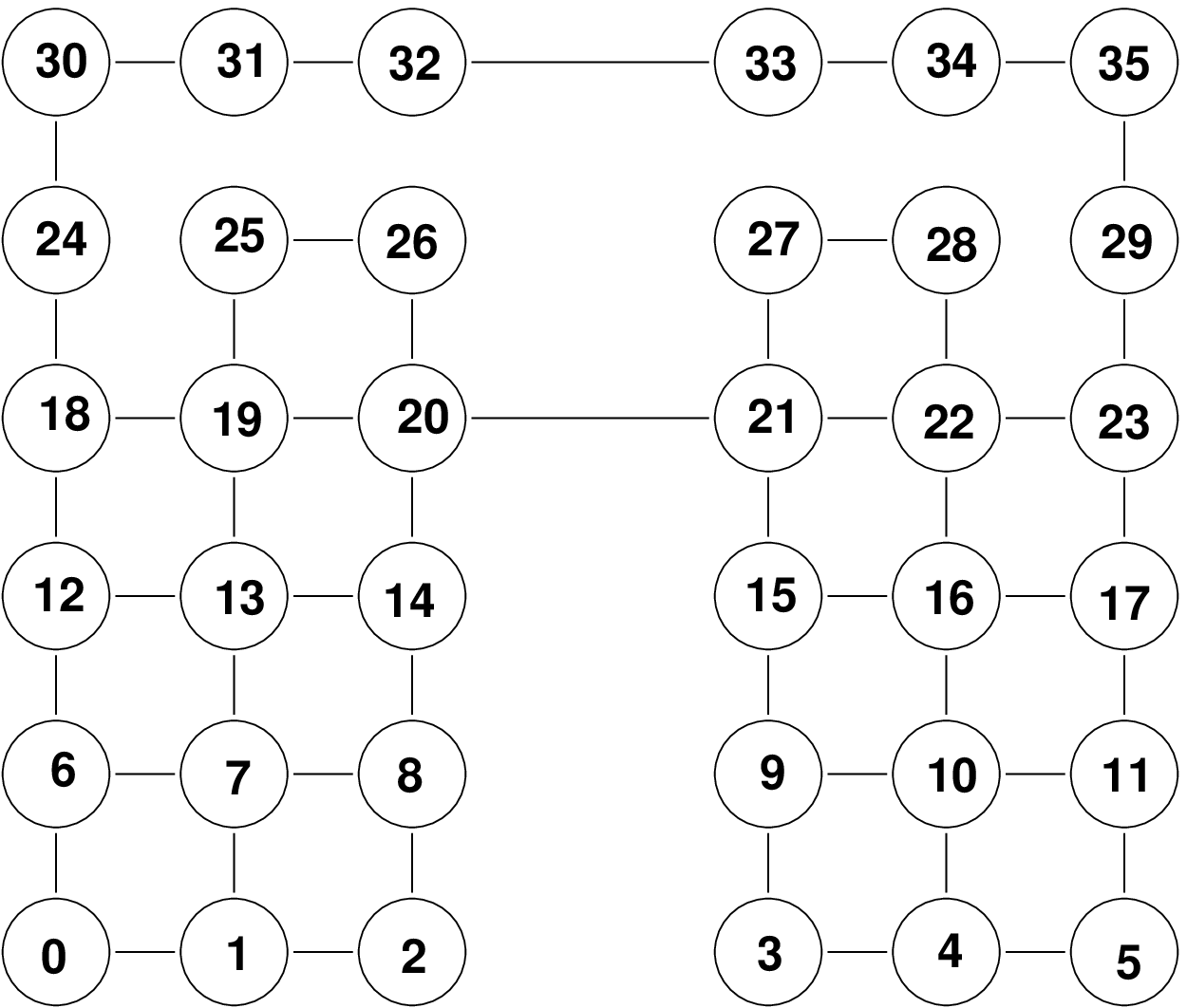,width=6cm}\hspace{3cm}\psfig{file=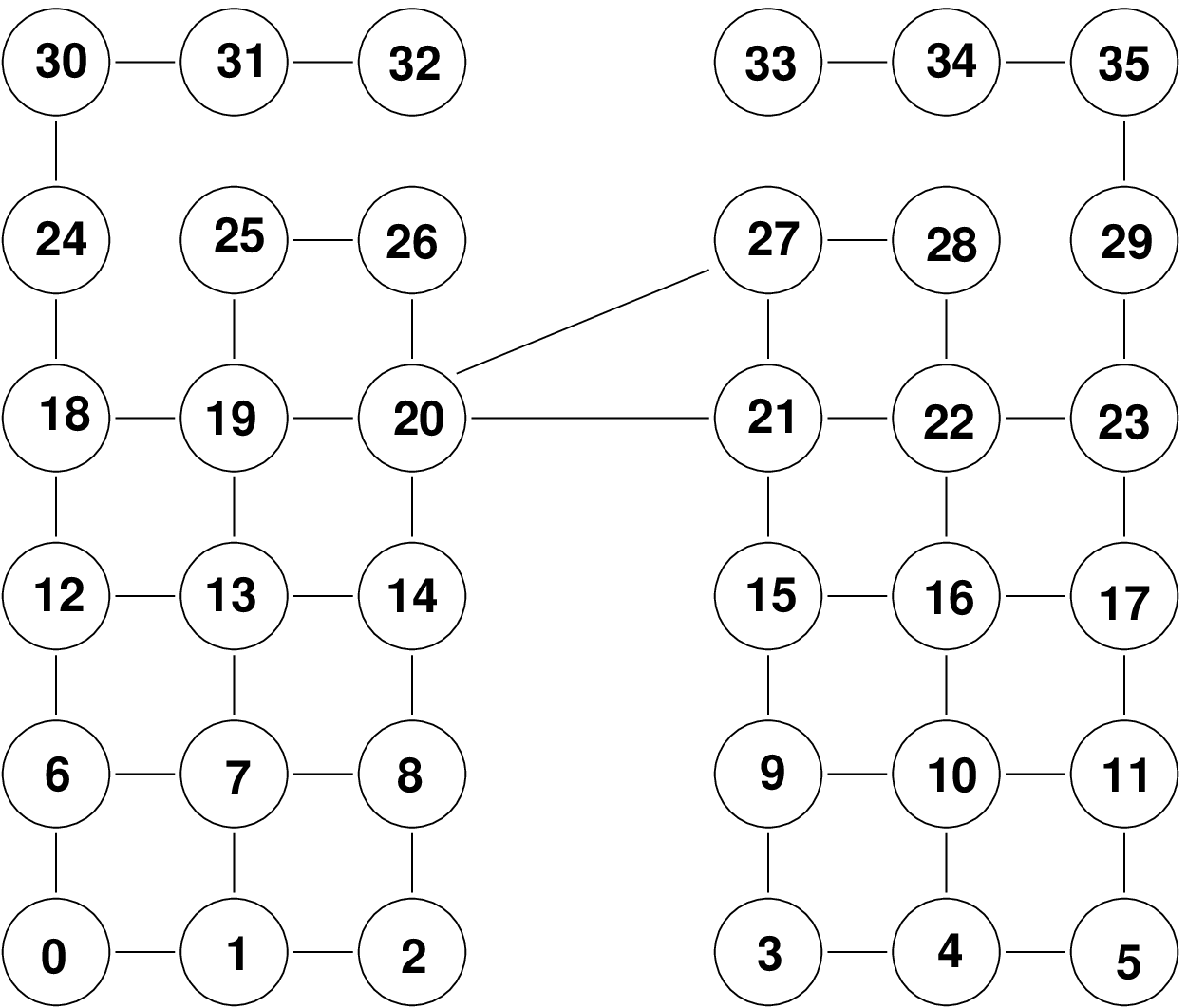,width=6cm}}
\caption{Left: Original 6x6 network. Right: Modified 6x6 network
favoring stochastic policies.}
\label{6x6net}
\end{figure*}

We test our algorithm on a domain adopted from Boyan and
Littman~\cite{Boyan94}. It is a discrete time simulator of
communication networks with various topologies and dynamic
structure. A communication network is an abstract representation of
real-life systems such as the Internet or a transport network. It
consists of a homogeneous set of nodes and edges between them
representing links (see Figure~\ref{6x6net}). Nodes linked to each
other are called neighbors. Links may be active ("up") or inactive
("down"). Each node can be the origin or the final destination of
packets, or serve as a router.

Packets are periodically introduced into the network with a uniformly
random node of origin and destination. They travel to their
destination node by hopping on intermediate nodes. No packets are
generated being destined to the node of origin. Sending a packet down
a link incurs a cost that could be thought of as time in
transition. There is an added cost to waiting in the queue of a
particular node in order to access the router's computational
resource---a queue delay. Both costs are assumed to be uniform
throughout the network. In our experiments, each is set to be a unit
cost. The level of network traffic is determined by the number of
packets in the network. Once a packet reaches its destination, it is
removed. If a packet has been traveling around the network for a long
time it is also removed as a hopeless case. Multiple packets line up
at nodes in an {\sc fifo} (first in first out) queue limited in
size. The node must forward the top packet in the {\sc fifo} queue to
one of its neighbors.

In the terminology of {\sc rl}, the network represents the environment whose
state is determined by the number and relative position of nodes, the
status of links between them and the dynamics of packets. The
destination of handled packets and the status of local links form the
node's observation. Each node is an agent who has a choice of
actions. It decides where to send the packet according to a
policy. The policy computed by our algorithm is stochastic, as opposed to deterministic, {\it
i.e.} it sends packets bound for the same destination down different links,
according to some distribution. The policy considered in our
experiments does not determine whether or not to accept a packet
(admission control), how many packets to accept from each neighbor, or
which packets should be assigned priority.

The node updates the parameters of its policy based on the reward.
The reward comes in the form of a signal distributed through the
network by acknowledgment packets once a packet has reached its final
destination. The reward depends on the total delivery time for the
packet. We measure the performance of the algorithm by the average
delivery time for packets once the system has settled on a policy
(ordinate axes on figure~\ref{res}). We apply policy shaping by
explicitly penalizing loops in the route. Each packet is assumed to
carry some elements of its routing history in addition to obvious
destination and origin information. They include the time when the
packet was generated, the time the packet last received attention from
some router, the trace of recently visited nodes and the number of
hops performed so far. In case a packet is detected to have spent too
much time in the network failing to reach its destination, such packet
is discarded and the network is penalized accordingly. Thus, a
defining factor in our simulation is weather the number of hops
performed by a packet is more than a total number of nodes in the
network.

\section{Algorithmic details}

Williams introduced the notion of policy search via gradient ascent
for reinforcement learning in his \reinf~
algorithm~\cite{Williams87,Williams92}, which was generalized to a
broader class of error criteria by Baird and
Moore~\cite{Baird,Baird99}. The general idea is to adjust parameters
in the direction of the empirically estimated gradient of the aggregate
reward.  We assume standard Markov decision process {\sc mdp}
setup~\cite{Kaelbling96}.  Let us consider the case of a single agent
interacting with a partially observable {\sc mdp} ({\sc pomdp}). The
agent's policy $\mu$ is a so-called reactive policy represented by a
lookup table with a value $\theta_{oa}$ for each observation-action
(destination/link) pair.  The policy defines the probability of an
action given past history as a continuous differentiable function of a
set of parameters $\theta$ according to a softmax rule, where $\Xi$ is
a temperature parameter: $\mu(a,o,\theta)\!=\!\Pr\left(a(t)\!=\!a \big|
o(t)\!=\!o, \theta\right) = \frac{\exp\left({\theta_{oa}/\Xi}\right)}
{\sum_{a'} \exp\left({\theta_{oa'}/\Xi}\right) } > 0$. This rule
assures that for any destination $o$ any link $a'$ available at the
node is sometimes chosen with some small probability dependent on the
temperature $\Xi$.

We denote by $H_t$ the set of all possible experience sequences
$h=\tuple{o(1)\!,a(1)\!,r(1)\!,\ldots\!,o(t)\!,a(t)\!,r(t)\!,o(t\!+\!1)\!}$
of length $t$.  In order to specify that some element is a part of the
history $h$ at time $\tau$, we write, for example, $r(\tau,\!h)$ and
$a(\tau,h)$ for the $\tau^{th}$ reward and action in the history
$h$. We will also use $h^\tau$ to denote a prefix of the sequence
$h\!\in\!H_t$ truncated at time $\tau\!\leq\!t\!:$
$h^\tau\!\stackrel{\rm def}{=}\!\langle
o(1),a(1),r(1),\ldots,o(\tau),a(\tau),r(\tau),o(\tau\!+\!1)\rangle$. 
The value of following a policy $\mu$ with parameters $\theta$ is the
expected cumulative discounted (by a factor of $\gamma \in [0,1)$)
reward that can be written as
\begin{eqnarray*}
V(\theta)=\sum_{t=1}^\infty \gamma^t \sum_{h \in H_t} \Pr(h \mid
\theta) r(t,h)\;.
\end{eqnarray*}

If we could calculate the derivative of $V(\theta)$ for each
$\theta_{oa}$, it would be possible to do an exact gradient ascent on
value $V()$ by making updates $ \Delta \theta_{oa} = \alpha
\frac{\partial}{\partial \theta_{oa}}V(\theta)$ for some step size
$\alpha$. Let us analyze the derivative for each weight $\theta_{oa}$,
\begin{eqnarray*}
\frac{\partial V(\theta)}{\partial \theta_{oa}}\!=\!&\sum_{t =
1}^\infty\!\gamma^t\!\sum_{h \in H_t}\!\left[r(t,h)\frac{\partial
\Pr(h \mid \theta)}{\partial \theta_{oa}}\right] \\  =&\sum_{t =
1}^\infty \gamma^t \sum_{h \in H_t} \Pr(h \mid \theta) r(t,h) \\
\times&\!\sum_{\tau=1}^t \frac{\partial \ln \Pr\left( a(\tau,h) \mid
h^{\tau-1}, \theta \right)} {\partial \theta_{oa}}\;.
\end{eqnarray*}

However, in the spirit of reinforcement learning, we assume no
knowledge of a world model that would allow the agent to calculate
$\Pr(h\!\mid\!\theta)$, so we must retreat to stochastic gradient
ascent instead. We sample from the distribution of histories by
interacting with the environment, and calculate during each trial an
estimate of the gradient, accumulating the quantities: $ \gamma^t
r(t,h) \sum_{\tau=1}^t \frac{\partial \ln
%\Pr\left(a(\tau,h) \mid h^{\tau-1}, \theta \right)} 
\mu\left(a,o, \theta \right)}  {\partial \theta_{oa}},   $
%\frac{\partial V}{\partial \theta_{oa}} \!\approx\!
for all $t$. For a particular policy architecture, this can be readily
translated into a gradient ascent algorithm guaranteed to converge to
a local optimum $\theta^*$ of $V(\theta)$.  Under our chosen policy
encoding we get:
\begin{eqnarray*}
\frac{\partial \ln\mu(a,o,\theta)}{\partial \theta_{o'a'}} \;=\;
\left\{
\begin{array}{ll}
 0 & {\rm if\ }o' \neq o,\\ -\frac{1}{\Xi}\mu\left(a',o,\theta\right)
& {\rm if\ }o'=o, a'\neq a,\\ \frac{1}{\Xi}\big[1
-\mu(a,o,\theta)\big] &{\rm if\ }o'=o, a'=a.\\
\end{array}\right.\\
\end{eqnarray*}
Applying this algorithm in a network of connected controllers
basically constitutes the algorithm of routing by distributed gradient
ascent policy search ({\sc gaps}).

We compare the performance of our distributed {\sc gaps} algorithm to
three others, as follows. 
``Best'' is a static routing scheme based
on the shortest path counting each link as a single unit of routing
cost. We include this algorithm because it provides the basis
for most  current industry routing heuristics~\cite{Bellman57,Dijkstra59}. 
``Bestload''performs routing according to the shortest path while
taking into account queue sizes at each node. It is close to the
theoretical optimum among deterministic routing algorithms even though
the actual best possible routing scheme requires not simply computing
the shortest path based on network loads, but also analyzing how loads
change in time according to routing decisions. Since calculating the
shortest path at every single step of the simulation would be
prohibitively costly in terms of computational resources, we
implemented ``Bestload'' by readjusting the routing policy only after
a notable change of loads in the network. We consider 50 successfully
delivered packets to constitute a notable load change.
Finally, ``Q-routing'' is a distributed {\sc rl} algorithm applied
specifically to this domain by Littman and Boyan~\cite{Boyan94}.
While our algorithm is stochastic and performs policy search,
Q-routing is a deterministic, value search algorithm. Note that our
implementation of the network routing simulation is based on the
software Littman and Boyan used to test Q-routing. Even so,the results
of our simulation of ``Q-routing'' and ``Best'' on the ``6x6'' network
differ slightly from Littman and Boyan's due to certain modifications
in traffic modeling conventions. For instance, we consider a packet
delivered and ready for removal only after it has passed through the
queue of the destination node and accessed its computational
resources, and not merely when the packet is successfully routed to
the destination node by an immediate neighbor, as in the original
simulation.

We undertake the comparison between {\sc gaps} and the aforementioned
algorithms with one important caveat. The {\sc gaps} algorithm
explores the class of stochastic policies while all other methods pick
deterministic routing policies. Consequently, it is natural to expect
{\sc gaps} to be superior for certain types of network topologies and
loads, where the optimal policy is stochastic. Later, we show that our
experiments confirm this expectation.

We implement  the distributed {\sc gaps} in {\sc pomdp}.  In
particular, we represent each router as a {\sc pomdp}, where the state
contains the sizes of all queues, destinations of all packets, state
of links (up or down); the environment state transition function is a
law of the dynamics of network traffic; an observation $o$ consists of
the destination of the packet; an action $a$ corresponds to sending
the packet down a link to an adjacent node; and finally, the reward
signal is the average number of packets delivered per unit of
time. Each agent is using a {\sc gaps} {\sc rl} algorithm to move
parameterization values down the gradient of the average reward. It
has been shown~\cite{Peshkin00} that an application of distributed
{\sc gaps} causes the system as a whole to converge to local optimum
under stationarity assumptions. This algorithm is essentially the one
described in chapter~3 and developed in chapter~5 of Peshkin's
dissertation~\cite{PeshkinPhD}.

Policies were initialized in two different ways: randomly and based on
shortest paths. We tried initialization with random policy uniformly
chosen over parameter space. With such initialization results are very
sensitive to the learning rate. High learning rate often causes the
network to stick to local optima in combined policy space, with very
poor performance. Low learning rate results in a slow
convergence. What constitutes high or low learning rate depends on the
specifics of each network and we did not find any satisfactory
heuristics to set it.  Obviously, such features as average number of
hops necessary to deliver a packet under the optimal policy as well as
learning speed crucially depend on the particular characteristics of
each network such as number of nodes, connectivity and modularity.

These considerations led us to a different way of initializing
controllers. Namely, we begin by computing shortest path and set
controllers to route most of the traffic down the shortest path, while
occasionally sending a packet to explore an alternative link. We call
this ``$\epsilon$-greedy routing''. In our experiments, $\epsilon$ is
set to $.01$. We believe that this parameter would not qualitatively
change the outcome of our experiments since it only influences
exploratory behaviour in the beginning.

The exploration capacity of the algorithm is regulated in a different
way as well. Both temperature and learning rate are simply kept
constant both for considerations of simplicity and for maintaining the
controllers' ability to adjust to changes in the network, such as
links failure. However, our experiments indicate that having a
schedule for reducing learning rate after a key initial period of
learning would improve performance. Alternatively, it would be
interesting to explore different learning rates for the routing
parameters on one hand, and the encoding of topological features on
the other.

\section{Empirical results}

We compared the routing algorithms on several networks with various
number of nodes and degrees of connectivity and modularity, including
$116$-node ``{\sc lata}'' telephone network. On all networks, the {\sc
gaps} algorithm performed comparably or better than other routing
algorithms. To illustrate the principal differences in the behavior of
algorithms and the key advantages of distributed {\sc gaps}, we
concentrate on the analysis of two routing problems on networks which
differ in a single link location.

\begin{figure*}[!hbt]
\centerline{\psfig{file=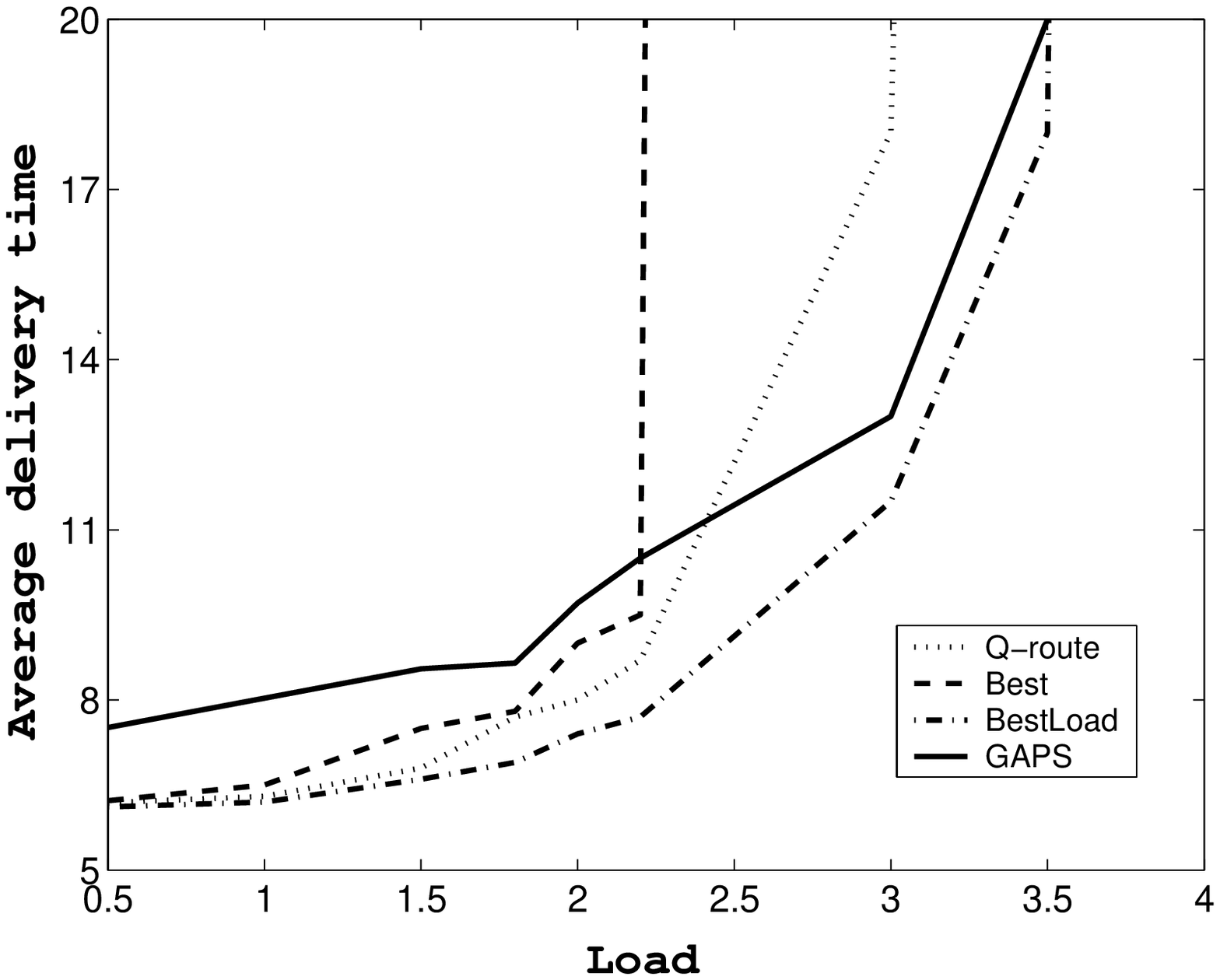,height=7cm}\hfill\psfig{file=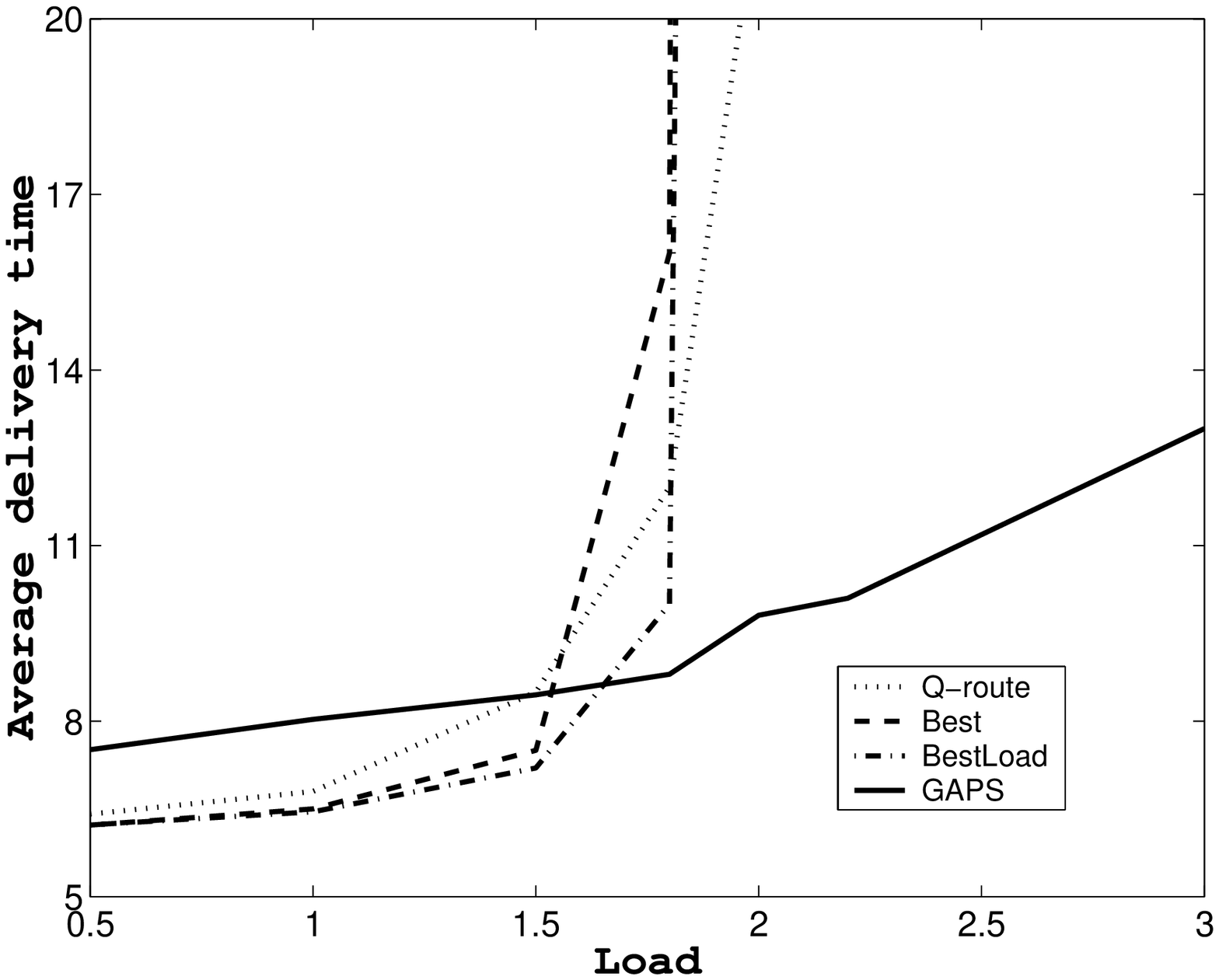,height=7.cm}}
\caption{Performance of routing algorithms on the original 6x6 network (left) and the modified 6x6 network (right).}
\label{res}
\end{figure*}

Figure~\ref{6x6net}.left presents the irregular 6x6 grid network
topology used by Boyan and Littman~\cite{Boyan94} in their
experiments. The network consists of two well connected components
with a bottleneck of traffic falling on two bridging links. The
resulting dependence of network performance on the load is depicted in
figure~\ref{res}.left. All graphs represent performance after the
policy has converged, averaged over five runs. We tested the network
on loads ranging from $.5$ to $3.5$, to compare with the results
obtained by Littman and Boyan. The load corresponds to the value of
the parameter of Poisson arrival process for the average number of
packets injected per time unit. On this network topology, {\sc gaps}
is slightly inferior to other algorithms on lower loads, but does at
least as well as Bestload on higher loads, outperforming both
Q-routing and Best. The slightly inferior performance on low loads is
due to exploratory behaviour of {\sc gaps} --- some fraction of
packets is always sent down random link.

To illustrate the difference between the algorithms more explicitly,
we altered the network by moving just one link from connecting nodes
$32$ and $33$, to connecting nodes $20$ and $27$ as illustrated by
figure~\ref{6x6net}.right.  Since node $20$ obviously represents a
bottleneck in this configuration,  the optimal routing policy is bound
to be stochastic. The resulting dependence of network performance on
the load is presented in figure~\ref{res}.right. {\sc gaps} is clearly
superior to other algorithms at high loads. It even outperforms
``Bestload'' that has all the global information in choosing a policy,
but is bound to deterministic policies. Notice how the deterministic
algorithms get frustrated at much lower loads in this network
configuration than in the previous one since from their perspective, the bridge
between highly connected components gets twice thinner (compare left
and right of Figure~\ref{res}).

The {\sc gaps} algorithm successfully adapts to changes in the network
configuration. Under increased load, the preferred route from the left
part of the network to the right becomes evenly split between the two
``bridges'' at node $20$. By using link $20-27$, the algorithm has to
pay a penalty of making a few extra hops compared to link $20-21$, but
as the size of the queue at node $21$ grows, this penalty becomes
negligible compared to the waiting time. Exploratory behavior helps
{\sc gaps} discover when links go down and adjust the policy
accordingly.  We have experimented with giving each router a few bits
of memory in finite state controller~\cite{PeshkinPhD,MeuleauUAI99}
but found that this does not improve performance and slows down the
learning somewhat.

\section{Related Work}

\nocite{Brown99} 

The application of machine learning techniques to the domain of
telecommunications is a rapidly growing area. The bulk of problems fit
into the category of resource allocation, {\it e.g.} bandwidth
allocation, network routing, call admission control ({\sc cac}) and
power management. {\sc rl} appears promising in attacking all of these
problems, separately or simultaneously.

Marbach, Mihatsch and Tsitsiklis~\cite{MarbachNIPS98} have applied an
actor-critic (value-search) algorithm to address resource allocation
within communication networks by tackling both routing and call
admission control. They adopt a decompositional approach,
representing the network as consisting of link processes, each with
its own differential reward.  Unfortunately, the empirical results
even on small networks, $4$ and $16$ nodes, show little advantage over
heuristic techniques.

Carlstr\"om~\cite{Carlstrom00} introduces another {\sc rl} strategy based on
decomposition called predictive gain scheduling. The control problem of
admission control is decomposed into a time-series prediction of
near-future call arrival rates and precomputation of control policies
for Poisson call arrival processes. This approach results in faster
learning without performance loss. Online convergence rate increases 50
times on a simulated link with capacity $24$~units/sec.

Generally speaking, value-search algorithms have been more extensively
investigated than policy search ones in the domain of
communications. Value-search (Q-learning) algorithms have arrived at
promising results. Boyan and Littman's~\cite{Boyan94} algorithm -
Q-routing, proves superior to non-adaptive techniques based on
shortest path, and robust with respect to dynamic variations in the
simulation on a variety of network topology, including an irregular $6
\times 6$ grid and 116-node {\sc lata} phone network. It regulates the
trade-off between the number of nodes a packet has to traverse and the
possibility of congestion.

Wolpert, Tumer and Frank~\cite{Wolpert98} construct a formalism for
the so-called Collective Intelligence ({\sc coin})neural net applied
to Internet traffic routing. The approach involves automatically
initializing and updating the local utility functions of individual
{\sc rl} agents (nodes) from the global utility and observed local
dynamics. Their simulation outperforms a Full Knowledge Shortest Path
Algorithm on a sample network of seven nodes. Coin networks employ a
method similar in spirit to the research presented here. They rely on
a distributed {\sc rl} algorithm that converges on local optima without
endowing each agent node with explicit knowledge of network
topology. However, {\sc coin} differs form our approach in requiring the
introduction of preliminary structure into the network by dividing it
into semi-autonomous neighborhoods that share a local utility function
and encourage cooperation. In contrast, all the nodes in our network
update their algorithms directly from the global reward.

The work presented in this paper focuses on packet routing using
policy search. It resembles the work of Tao, Baxter and
Weaver~\cite{TaoICML01} who apply a policy-gradient algorithm to
induce cooperation among the nodes of a packet switched network in
order to minimize the average packet delay. While their algorithm
performs well in several network types, it takes many (tens of
millions) trials to converge on a network of just a few nodes.

Applying reinforcement learning to communication often involves
optimizing performance with respect to multiple criteria. For a recent
discussion on this challenging issue see Shelton~\cite{SheltonPhD}. In
the context of wireless communication it was addressed by
Brown~\cite{Brown00} who considers the problem of finding a power
management policy that simultaneously maximizes the revenue earned by
providing communication while minimizing battery usage. The problem is
defined as a stochastic shortest path with discounted infinite
horizon, where discount factor varies to model power loss. This
approach resulted in significant ($50\%$) improvement in power usage.

Gelenbe et al.~\cite{Gelenbe01} also compute the reward as a weighted
combination of the probability of packet loss and packet delay. The
packets themselves are agents controlling routing and flow control in
a Cognitive Packet Network. They split packets into three types:
"smart", "dumb" and "acknowledgment". A small number of smart packets
learn the most efficient ways of navigating through the network, dumb
packets simply follow the route taken by the smart packets, while
acknowledgment packets travel on the inverse route of smart packets to
provide source routing information to dumb packets. The division
between smart and dumb packet is an explicit representation of the
explore/exploit dilemma. Smart packet allow the network to adapt to
structural changes while the dumb packets exploit the relative
stability between those changes. Promising results are obtained both
on a simulation network of $100$ nodes and on a physical network of $6$
computers.

Subramanian, Druschel and Chen~\cite{Subramanian97} adopt an approach from
ant colonies that is very similar in spirit. The individual hosts in their
network keep routing tables with the associated costs of sending a packet to
other hosts (such as which routers it has to traverse and how expensive they
are). These tables are periodically updated by "ants"-messages whose function
is to assess the cost of traversing links between hosts. The ants are
directed probabilistically along available paths. They inform the hosts along
the way of the costs associated with their travel. The hosts use this
information to alter their routing tables according to an update rule. There
are two types of ants. Regular ants use the routing tables of the hosts to
alter the probability of being directed along a certain path. After a number
of trials, all regular ants on the same mission start using the same routes.
Their function is to allow the host tables to converge on the correct cost
figure in case the network is stable. Uniform ants take any path with equal
probability. They are the ones who continue exploring the network and assure
successful adaptation to changes in link status or link cost.

\section{Discussion}

Admittedly, the simulation of network routing process presented here
is far from being realistic. A more realistic model could include such
factors as non-homogeneous networks with regard to links bandwidth and
routing nodes buffer size limits, collisions of packets, packet
ordering constraints, various costs associated with say, particular
links chosen from commercial versus government subnetworks, minimal
Quality of Service requirements. Introducing priorities for individual
packets brings up yet another set of optimization issues.
However, the learning algorithm we applied shows promise in handling
adaptive telecommunication protocol and there are several obvious ways
to develop this research. Incorporating domain knowledge into
controller structure is one such direction. It would involve
classifying nodes into sub-networks and routing packets in a
hierarchical fashion.
One step further down this line is employing learning algorithms for
routing in ad-hoc networks. Ad-hoc networks are networks where nodes
are being dynamically introduced and terminated from the system, as
well as existing active nodes are moving about, loosing some
connections and establishing new ones. Under the realistic assumption
that physical variations is the network are slower than traffic routing
and evolution, adaptive routing protocol should definitely outperform
any heuristic pre-defined routines. We are currently pursuing this
line of research.

\bibliography{bib/lpk,bib/thesis}
\end{document}